\documentclass[conference]{IEEEtran}
\IEEEoverridecommandlockouts
\usepackage{cite}
\usepackage{amsmath,amssymb,amsfonts}
\usepackage{graphicx}
\usepackage{textcomp}
\usepackage{xcolor}
\usepackage{ulem}
\usepackage{subfigure}

\usepackage{enumitem,amssymb}
\usepackage{hyperref}
\newlist{todolist}{itemize}{2}
\setlist[todolist]{label=$\square$}
\usepackage{pifont}
\def\BibTeX{{\rm B\kern-.05em{\sc i\kern-.025em b}\kern-.08em
    T\kern-.1667em\lower.7ex\hbox{E}\kern-.125emX}}
\begin{document}

\title{Convolutional Neural Network Modelling for MODIS Land Surface Temperature Super-Resolution \\
%{\footnotesize \textsuperscript{*}Note: Sub-titles are not captured in Xplore and should not be used}
%\thanks{This work was supported by }
}

\author{\IEEEauthorblockN{Binh Minh Nguyen\IEEEauthorrefmark{1}\IEEEauthorrefmark{5}, Ganglin Tian\IEEEauthorrefmark{1}\IEEEauthorrefmark{5}, Minh-Triet Vo\IEEEauthorrefmark{1}\IEEEauthorrefmark{5}, Aur\'elie Michel\IEEEauthorrefmark{2}, Thomas Corpetti\IEEEauthorrefmark{3} \\ and Carlos Granero-Belinchon\IEEEauthorrefmark{1}\IEEEauthorrefmark{4}.}
\IEEEauthorblockA{\IEEEauthorrefmark{1}\textit{Mathematical and Electrical Engineering Department}, 
\textit{IMT Atlantique, Lab-STICC, UMR CNRS 6285},
F-29238 Brest, France}
%carlos.granero-belinchon@imt-atlantique.fr}
\IEEEauthorblockA{\IEEEauthorrefmark{2}\textit{ONERA-DOTA},
\textit{University of Toulouse},
F-31055 Toulouse, France}
\IEEEauthorblockA{\IEEEauthorrefmark{3}\textit{CNRS},
\textit{UMR 6554 LETG},
F-35043 Rennes, France}
\IEEEauthorblockA{\IEEEauthorrefmark{4}Corresponding author: carlos.granero-belinchon@imt-atlantique.fr}
\IEEEauthorblockA{\IEEEauthorrefmark{5} These authors contributed equally to this work.}}
\maketitle

\begin{abstract}

Nowadays, thermal infrared satellite remote sensors enable to extract very interesting information at large scale, in particular Land Surface Temperature (LST). However such data are limited in spatial and/or temporal resolutions which prevents from an analysis at fine scales. For example,  MODIS satellite  provides daily acquisitions with 1Km spatial resolutions  which is not sufficient  to deal with highly heterogeneous environments as agricultural parcels.
Therefore, image super-resolution is a crucial task to better exploit MODIS LSTs. This issue is tackled in this paper. We introduce a deep learning-based algorithm, named Multi-residual U-Net, for super-resolution of MODIS LST single-images. Our proposed network is a modified version of U-Net architecture, which aims at super-resolving the input LST image 
%from 4 to 1Km per pixel and 
from 1Km to 250m per pixel. The results show that our Multi-residual U-Net outperforms other state-of-the-art methods.
\end{abstract}


\section{Introduction}
Land Surface Temperature (LST) images at a high temporal resolution are of prime importance to efficiently monitor physical processes related to climate change such as water stress, evapotranspiration, wildfires or urban heat islands~\cite{sun2008retrieval,Kerr2000}. LST is retrieved from remote sensing images in the Thermal InfraRed (TIR) spectral domain. Thus, sensors such as SEVIRI (15 minutes
revisit time) or MODIS (12h revisit time) present interesting temporal resolutions. However, these sensors do not generally retrieve the LST at a satisfactory spatial resolution for local scale applications and fine scale analysis, especially for highly heterogeneous environments like urban areas, diverse agricultural plots or sparse forests.

For this reason, super-resolution methods are required to improve the native resolution of sensors. While in the reflective domain of the electromagnetic spectrum both statistical and Artificial Intelligence (AI) methods of super-resolution have been developed~\cite{brodu2017super},  in the TIR domain only classical statistical methods have been applied at the moment~\cite{agam2007vegetation, wang2016area, granero2019multi}. Though efficient, these approaches lead to limitations such as 1) the needing of high resolution products in the Visible and Near-Infrared (VNIR) or Short Wave Infrared (SWIR) domains acquired in the same area and close in time and 2) scale-invariant hypotheses which sometimes are not adapted. 

This work aims to develop Convolutional Neural-Network (CNN) models for single MODIS LST image super-resolution. We introduce \textbf{Multi-residual U-Net},  developed based on the U-Net architecture~\cite{10.1007/978-3-319-24574-4_28}, that can efficiently tackle the task of LST super-resolution. This is a promising result as no operational methods exist yet. 
In next section, data are introduced. At first, we describe MODIS data  used  for training and validation purposes (both training and validation are performed when upscaling LST from 4Km to 1Km of spatial resolution). A final validation of MODIS super-resolution products from 1Km to 250m is done by comparing them to ASTER LST data of the same area and acquired close in time.  In section \ref{sec:NN}, the neural network approach is presented. This technique does not require  VNIR-SWIR high resolution products, however it is still submitted to scale-invariant hypotheses due to the differences in the range of scales used during training (4Km-1Km) and in final applications (1Km-250m).  Lastly, experimental results and discussions are presented in section \ref{sec:RES}.

\section{MODIS and ASTER data}\label{sec:data}

\subsection{MODIS}
\begin{figure}[htbp]
    \centering
    \includegraphics[width=\linewidth]{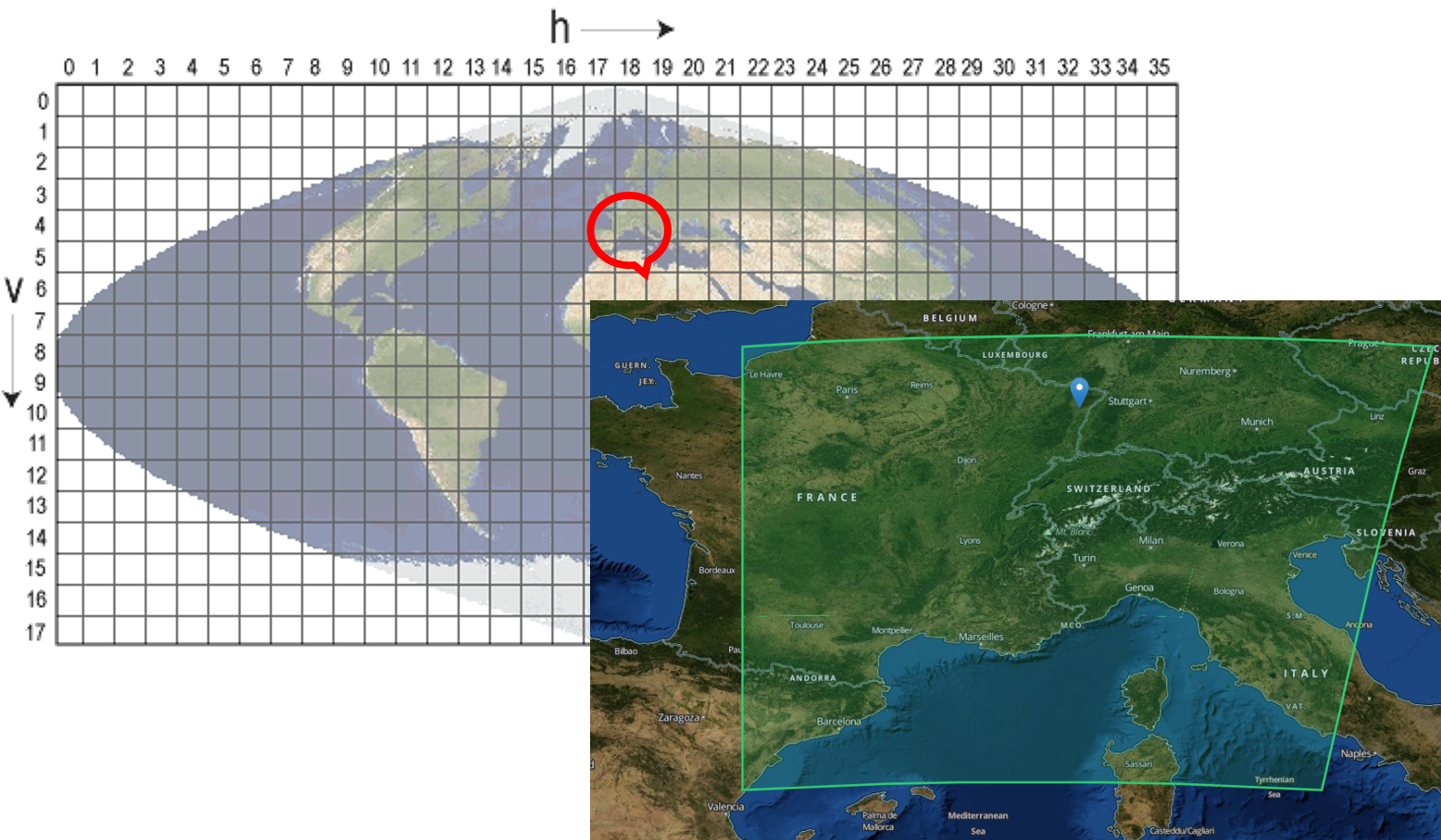}
    \caption{Center Europe area (h18v04 MODIS tile) used in this study. The blue marker indicates the location of Strasbourg at the border between France and Germany.}
    \label{Fig:h18v04 tile used for validation}
\end{figure}

MODIS is a NASA instrument devoted to the land, atmosphere and oceans observation and is widely used for climate change studies. MODIS has 36 bands ranging from the visible to the TIR domain (0.459 $\mu$m-14.385 $\mu$m). Bands corresponding to the VNIR domain present spatial resolutions of 250m, 500m and 1000m depending on the band. Bands corresponding to SWIR and TIR domains present a spatial resolution of 1Km~\cite{modisweb}. In this paper, we focus on the MOD11A1 MODIS Land Surface Temperature (LST) product~\cite{wanmod11a1}. Specifically, MOD11A1 provides LST at a resolution of 1Km, with a temporal resolution of days, with tiles of 1200 $\times$ 1200Km. We evaluate our model on areas located in the center of Europe which correspond to h18v04 tile under the MODIS geographic distribution standard, as shown in Fig~\ref{Fig:h18v04 tile used for validation}. 

We use 4 years of daily MODIS LST data from 2017 to 2020 as training dataset, with a total number of 2920 images corresponding to 1460 different dates, each one with daytime and nighttime data. Accordingly, we use 4083 images from 2011 to 2013 as validation set. Then these images of size 1200 by 1200 pixels are sliced to obtain small ones of size 64 by 64 pixel, which will be easier to drive by our models.

It should be noted that one of the main limitations for collecting LST data is the cloud cover, leading to missing pixels in the images. In the last years, several algorithms have been developed to reconstruct the missing pixel values by using auxiliary data such as Vegetation Index\cite{zeng2018two}, elevation\cite{kilibarda2014spatio} and Land Cover (LC) \cite{fan2014reconstruction}. However,  these methods introduce reconstruction errors and increase the uncertainty of the model prediction results. Therefore, we do not reconstruct missing pixels, but simply discard patches containing clouds, resulting in a total of 21650 images of 64 by 64 pixel without missing pixels. 

\subsection{ASTER} \label{sec:ASTER}

ASTER is a multispectral mission that includes 14 spectral bands from the visible to the TIR domain (0.52$\mu$m-11.65$\mu$m), and with spatial resolutions of 15m (VNIR), 30m (SWIR) and 90m (TIR), all with a swath width of 60Km. 

In this paper we use the ASTER LST product as the best reference for the validation of our MODIS super-resolved images \cite{asterweb}. For validation purposes and due to the low revisit time of ASTER, we focus on MODIS and ASTER concomitant images acquired at the French-German border close to Strasbourg (blue marker in figure~\ref{Fig:h18v04 tile used for validation}) on the 26$^{th}$ January 2017. As before, we retain only cloud-free patches of 64Km by 64Km. Lastly, for comparison and validation purposes, both products have the same map projection. Thus, MODIS products which are provided in a sinudoidal projection are converted into the UTM projection using the GDAL library in Python, version $3.6$.

\section{Super-resolution methods for LST}\label{sec:NN}

\subsection{Classical machine learning methods}
In the TIR domain, a family of statistical sharpening techniques namely TsHARP~\cite{agam2007vegetation} and ATPRK~\cite{wang2016area} have been developed to solve the image super-resolution task. These methods rely on two main hypotheses: 1) the relationship between the LST and VNIR-SWIR features such as the Normalized Difference Vegetation Index (NDVI) is linear and 2) this relationship is scale invariant. In other word, based on these hypotheses, it is possible to obtain high quality resolution LST images by modelling the relationship between LST and VNIR-SWIR features at a coarse scale, and applying the model to high resolution VNIR-SWIR features to retrieve the LST images at this finer scale. Finally, in order to enhance the high resolution result, a residual estimation is applied to correct  fine-scale LST values. A review of these statistical sharpening methods can be found in~\cite{granero2019multi}.

Since ATPRK presented the best performances in~\cite{granero2019multi}, we chose it as a representative method from the classical machine learning group to benchmark against our results from CNN-based approach.

\subsection{Neural Network approach}
U-Net was introduced in 2015 by Ronneberger et al.~\cite{10.1007/978-3-319-24574-4_28} and has been applied widely for semantic segmentation task. Moreover, it is a universal network which can be modified and employed in number of remote sensing applications such as pansharpening~\cite{YAO2018364} or even road extraction~\cite{8309343}. The key element which makes U-Net robust is the idea of using long skip connection on its symmetric network structure. This allows the features extracted from the encoder to propagate to the decoder, giving an information flow from low to high network levels. By this way, the architecture of U-Net not only enhances the construction path of the output from the high-level feature, but also reduces significantly the problem of gradient vanishing.

\subsubsection{Multi-residual U-Net}
\begin{figure*}[htp]
    \centering
    \includegraphics[width=\linewidth]{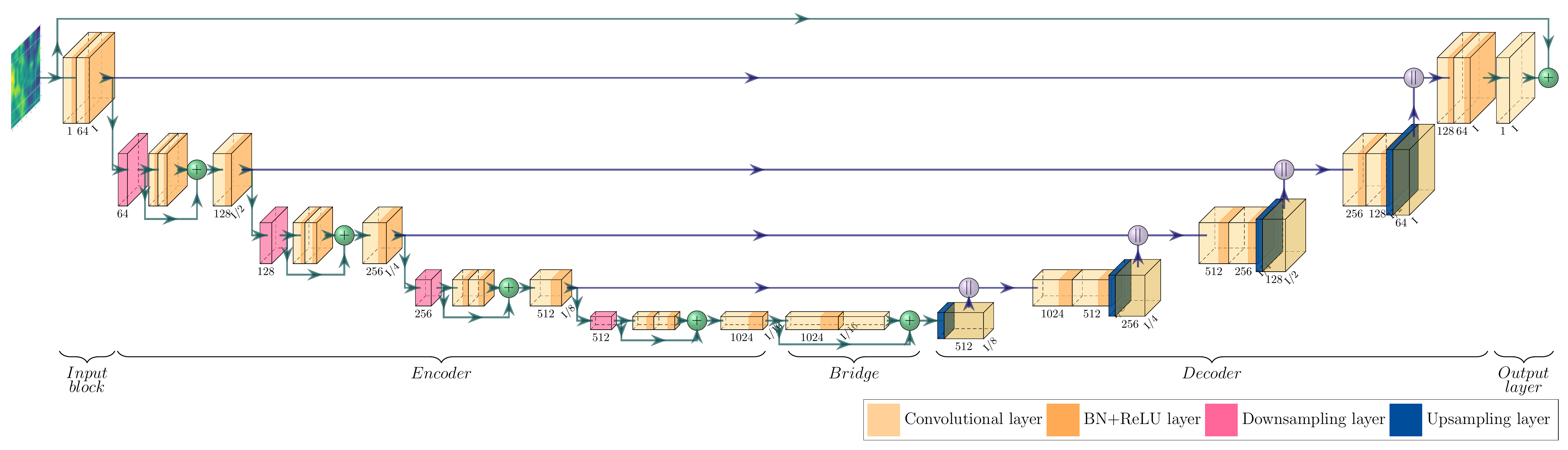}
    \caption{Multi-residual U-Net architecture outline. The green circle with a "+" symbol represents a sum operation. The purple one filled with a "$||$" symbol along with purple arrows define a concatenation operation between the output of previous upsampling layer and the corresponding encoder layer.}
    \label{Fig:Our model}
\end{figure*}

The architecture of Multi-residual U-Net is illustrated in Figure~\ref{Fig:Our model}. It can be decomposed into the input block, the encoder, the bridge, the decoder and the output layer. The input block and the encoder extract the input's features from low to high level;  the decoder reconstructs the output from these encoded features. The ``bridge'', in  the middle part of the architecture, lies between the encoder and the decoder. Comparing with the original U-Net, the proposed \textbf{Multi-residual U-Net} has three main differences located in the input/output layers, the encoder and the bridge. 
First, the network takes a bicubic-Interpolated Low Resolution (ILR) image as input, and maps it to a residual image, which is difference between the desired final high resolution (HR) output and the ILR input. This end-to-end residual mapping is widely employed in single image super-resolution networks such as VDSR~\cite{kim2016accurate} and DMCN~\cite{8518855}, to ease the learning process. 
Secondly, our proposed architecture uses a convolutional layer, instead of a max-pooling one, to downsample the feature map. Although downsampling by convolution increases the number of parameters in the network, it retains more information of its input feature maps. Thus the convolution operation takes into account all of the neuron values in its perceptive field to produce the output, not just the maximum value as in max-pooling downsampling. 
Furthermore, the fully convolutional encoder of U-Net is replaced by our residual-style encoder, which is inspired by the work of He et al.~\cite{he2015deep}. Following them, we use residual learning to effectively resolve the degradation problem of deep network caused by gradient vanishing/exploding, and so we guarantee its performance. Based on their structure of the residual building block, we proposed a replacement of two consecutive convolutional layers in the original U-Net architecture by one residual unit followed by a convolution block (sequentially made of a convolutional layer and a batch normalization layer followed by the ReLU activation). A residual unit includes two consecutive convolution blocks with the same number of filters, along with an identity mapping~\cite{he2015deep}. 
The final modification in \textbf{Multi-residual U-Net} is the bridge. Our approach is inspired by Johnson et al.~\cite{johnson2016perceptual} who placed between the encoder and the decoder a transformation structure composed of several residual blocks, which are the same building blocks of ResNet introduced in~\cite{he2015deep}. According to~\cite{johnson2016perceptual}, this transformation structure plays the role of a connection, which transforms the features at the last encoding block to the first block of the decoding path, increasing the performance of their generator network in style transfer and super-resolution objectives. From experiments, the best performances of our \textbf{Multi-residual U-Net} are found with only one residual block, and so we reduce the number of blocks in the bridge of our model to one.

\subsubsection{Loss function}

We used Mean Squared Error (MSE) as the principle loss function during the training process. MSE is frequently used in Neural Network training since it has been proved to provide good performances~\cite{dong2015image}.

\begin{equation}\label{MSE loss}
	L_{MSE} = \frac{1}{N.M}\sum_{i=1}^{N}\sum_{j=1}^{M} (LST^{GT}_{i,j} - LST^{SR}_{i,j})^2
\end{equation}

where $LST^{GT}_{i,j}$ is the LST of the $ j^{th} $ pixel of $ i^{th} $ high resolution ground truth image and $LST^{SR}_{i,j}$ is the corresponding LST in the super-resolution output of the deep learning model. $N$ is the number of images in a mini-batch, and $M$ is the total number of pixels within an image.

\subsubsection{Training setting}
To train our \textbf{Multi-residual U-Net}, we first split the 4-year MODIS LST dataset from 2017 to 2020 into train and test set with 75\% and 25\% of images respectively with daytime and nighttime jumbled. 
The inputs of the neural network were the ILR images, which were obtained by first downsampling with Norm-L4, according to Stefan-Boltzmann law, each $64 \times 64$ image at 1Km of spatial resolution to one at 4Km of spatial resolution, and second by upscaling these images using bicubic interpolation to return newly an image with size $64 \times 64$ and spatial resolution of 1Km. 
Before feeding into the model, these ILR images were divided by the maximum value from the train set for normalization. 

So, MODIS $64 \times 64$ LST images at 1Km of spatial resolution are used as ground truth during training, while normalized ILR images of same dimensions and spatial resolution are used as inputs. We trained our model for 300 epochs with a batch size of 32. During the training process, we set the learning rate at 0.0001 for the first 50 epochs and divided it by 100 for the rest of the process.

\section{Results and discussion}\label{sec:RES}

\subsection{Performance Assessment}

To evaluate the performances of our model on single LST image super-resolution compared to other available methods, we employed the standard RMSE (Root Mean Square Error) defined as the root square of~eq.(\ref{MSE loss}). Furthermore, we used PSNR (Peak Signal-to-Noise Ratio) and SSIM (Structure Similarity Index)~\cite{1284395} which are well-known metrics for quantifying the resemblance between two images:

\begin{equation}
    \begin{aligned}
        \operatorname{PSNR} = 20\log_{10}\left(\frac{DR_{GT}}{\sqrt{MSE}}\right)
    \end{aligned}
\end{equation}

\begin{equation}
    \begin{aligned}
        \operatorname{SSIM}=\frac{\left(2 \mu_{GT} \mu_{SR}+c_{1}\right)\left(2 \sigma_{GT SR}+c_{2}\right)}{\left(\mu_{GT}^{2}+\mu_{SR}^{2}+c_{1}\right)\left(\sigma_{GT}^{2}+\sigma_{SR}^{2}+c_{2}\right)}
    \end{aligned}
\end{equation}
where the subscripts $GT$ and $SR$ denote the high resolution ground truth image and the super-resolution model's output. The term $DR_{GT}=MAX_{GT}-MIN_{GT}$ is the dynamic range, measured by the difference between the highest pixel value and the lowest pixel value of the ground truth image; $MSE$ is the mean-square error between $GT$ and $SR$, see eq.(\ref{MSE loss});  $\mu$ represents the mean of the image, $\sigma$ represents the variance of the image (in case the subscript is $GT$ or $SR$) or the covariance of two images (in case the subscript is $GTSR$) and $c_{1}=(0.01 \cdot DR_{GT})^2$ and $c_{2}=(0.03 \cdot DR_{GT})^2$ are two stabilizing factors in SSIM computation. 

\subsection{Results on validation dataset}
\begin{figure}[htp]
    \centering
    \includegraphics[width=0.45\textwidth]{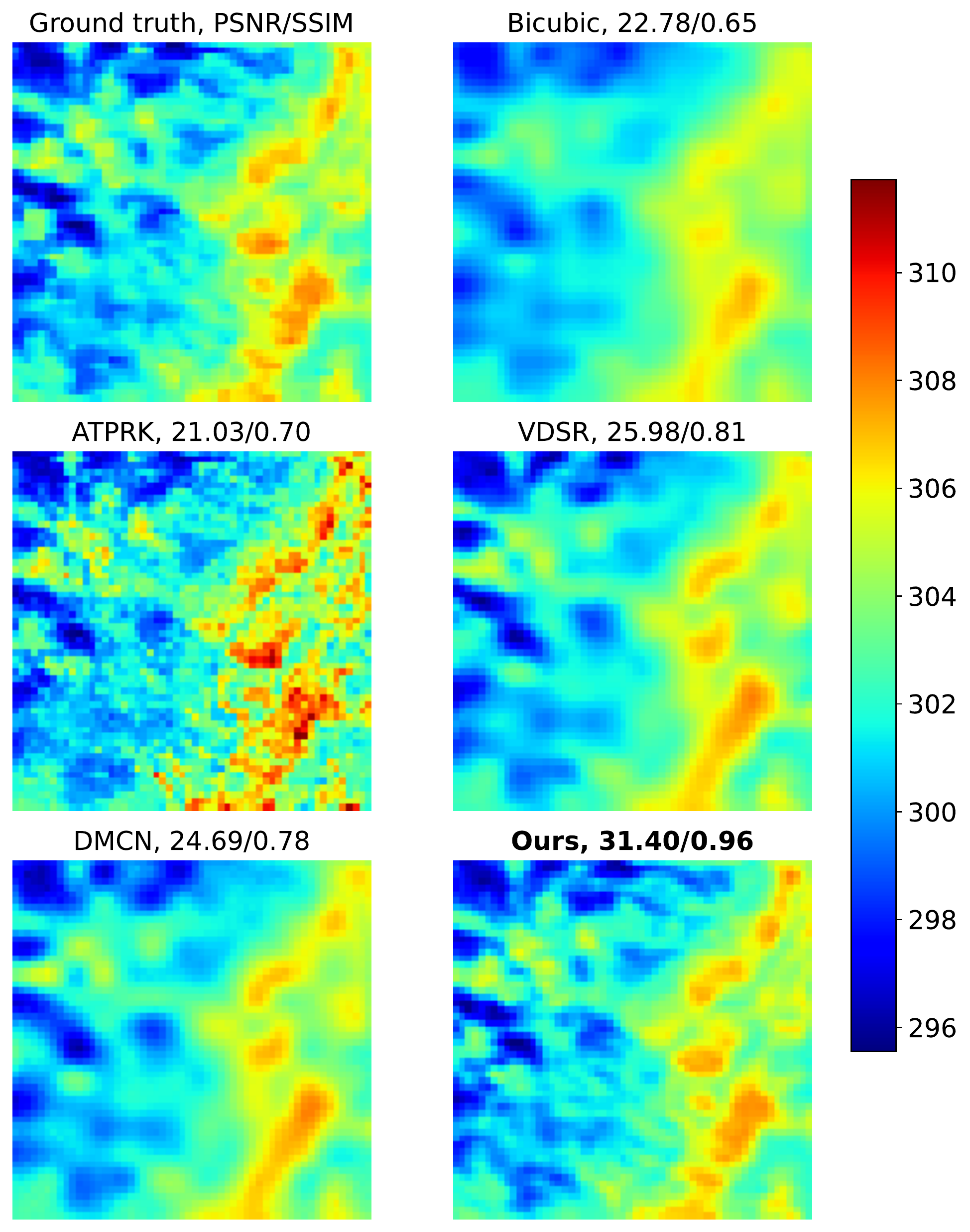}
    \caption{Qualitative results of super-resolution methods with scale ratio $\times 4$, here going from 4 to 1Km of resolution.}
    \label{Fig:visualization}
\end{figure}

Since we trained our models on 4-year LST dataset from 2017 to 2020, we used images from 2011-2013 to evaluate the performances. Within the validation set, each LST image is paired with a NDVI image in the same region and with the closest acquisition time. The pairs are served as the input of ATPRK method. For other neural network approaches, only single LST images are required. The benchmark results are shown in the table \ref{tab:result} and the qualitative comparison is demonstrated in Figure \ref{Fig:visualization}.

\begin{table}[!ht]
    \centering
    \caption{Evaluation of several methods for LST image super-resolution with scale ratio $\times 4$. The bold numbers indicate the best performance. }
    \resizebox{0.85\linewidth}{!}{
    \begin{tabular}{|l|l|l|l|}
    \hline
        Method & PSNR & SSIM & RMSE \\ \hline \hline
        Bicubic & 23.91 & 0.61 & 0.69 \\ 
        ATPRK & 21.59 & 0.61 & 0.90\\  
        VDSR & 25.42 & 0.72 & 0.58\\  
        DCMN & 25.05 & 0.71 & 0.61\\
        \bf{Multi-residual U-Net} & \bf{28.40} & \bf{0.85} & \bf{0.39} \\ \hline 
    \end{tabular}}
    \label{tab:result}
\end{table}

As can be shown in table \ref{tab:result}, our \textbf{Multi-residual U-Net} provided the best performance among considered approaches. Thanks to a complex architecture, our model surpassed other methods with a significant improvement in PSNR and SSIM. As shown  in Figure \ref{Fig:visualization}, it recovered a much more detailed image than VDSR and DMCN. The output of \textbf{Multi-residual U-Net} had high similarity compared to the high resolution ground truth. Besides, the classical method ATPRK recovered well the overall structure of the ground truth and provided an impression of a high contrast image thanks to the relationship with NDVI. However much information was unable to be brought back correctly explaining the low PSNR value.

Although our model outperformed other techniques, especially bicubic, there were certain cases in which bicubic achieved better performances. The major reason came from low dynamic ranges within LST small images. When dealing with low variability, linear interpolation methods such as bicubic can provide a good approximation with little amount of error since the distances between values are relatively small. Nevertheless, in this evaluation, only 34 out of 4083 images (which means less than 1\%) fit the mentioned circumstance. These cases are minorities and have tiny impact on quantitative performance assessment.

\subsection{Validation with ASTER}
The main objective of this research is to investigate the possibility of improving the resolution from 1Km to 250m per pixel, which is the best spatial resolution of the VNIR MODIS images. To evaluate the performance of our model for this goal, we employed ASTER data mentioned in section \ref{sec:ASTER} as the reference. 

Since the ASTER LST images present an initial spatial resolution of 90m, we use an area-weighted linear downsampling to obtain ASTER LST images at 250m of resolution, i.e. a linear downsampling which takes into account the considered area of the pixels used in the interpolation.

\begin{figure}[htp]
    \centering
    \includegraphics[width=0.4\textwidth]{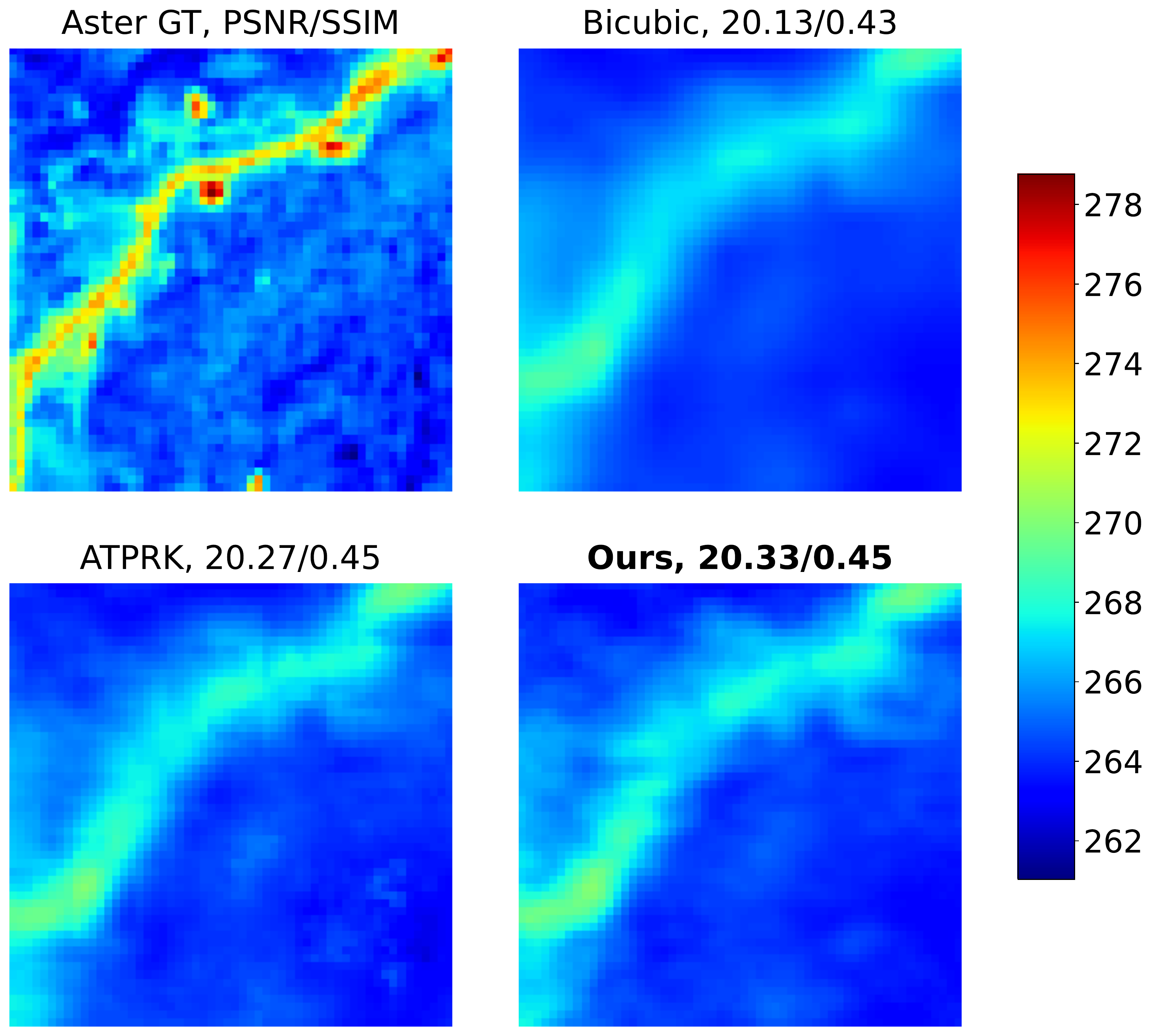}
    \caption{Qualitative results of super-resolution methods with scale ratio $\times 4$, here going from 1Km to 250m of resolution.}
    \label{Fig:visualization_aster}
\end{figure}

In this section, we compare ours \textbf{Multi-residual U-net}, representing neural network based approaches, with ATPRK and bicubic. The quantitative evaluation is done in table \ref{tab:result_aster} and the visualization is in figure \ref{Fig:visualization_aster}.

\begin{table}[!ht]
    \centering
    \caption{Quantitative comparison between ASTER and Super-Resolution MODIS at 250m for 7 cloud-free images.}
    \resizebox{0.85\linewidth}{!}{
    \begin{tabular}{|l|l|l|l|}
    \hline
        Method & PSNR & SSIM & RMSE \\ \hline \hline
        Bicubic & 19.56 & 0.39 & 1.48 \\   
        ATPRK & 19.66 & 0.41 & 1.46\\ 
        \bf{Multi-residual U-Net} & \bf{19.68} & \bf{0.42} & \bf{1.46} \\ \hline 
    \end{tabular}}
    \label{tab:result_aster}
\end{table}

The reconstruction results are not pleasing visually since our model attempted to recover the resolution of 250m from 1Km per pixel with the learned parameters trained when going from 4Km to 1Km per pixel. So, this indicates that scale-invariant hypotheses of classical machine learning methods are not adapted in these ranges. The use of a local loss (as gradient loss of VGG loss) would probably help in recovering finer structures. Nevertheless, it should be outlined that the presented metrics of \textbf{Multi-residual U-Net} shows slight improvement compared to bicubic and similar performances than ATPRK, which indicates the possibility of appearance of repeatable characteristics occurring in both upscaling process.% (from 4Km to 1Km and from 1Km to 250m).

\section{Conclusions}
In this paper, we proposed a new U-net based architecture called \textbf{Multi-residual U-Net} to solve the super-resolution task for MODIS LST images from 1Km to 250m per pixel (code and model are freely available at \href{https://github.com/IMT-Project-LTS-SR/MRUNet-for-MODIS-super-resolution}{https://github.com/IMT-Project-LTS-SR/MRUNet-for-MODIS-super-resolution}). During training on MODIS data from 4Km to 1Km per pixel, the proposed architecture outperformed other neural network approaches such as VDSR and DMCN as well as ATPRK. Moreover, \textbf{Multi-residual U-Net} applied on MODIS LST from 1Km to 250m per pixel reached the performances of ATPRK. Future research will deal with compensating the scale-invariant hypothesis which is not adapted for resolutions ranging from 4Km to 250m. Finally, we point out the interest of such a super-resolution method for the future mission TRISHNA~\cite{8518720}, since these methods apply on single LST images without needing VNIR-SWIR products at better resolutions.

\bibliographystyle{IEEEtran} 
\bibliography{biblio}

\end{document}